\documentclass{article} % For LaTeX2e
\usepackage{iclr2025_conference,times}

\usepackage{amsmath,amsfonts,bm}

\def\eqref#1{equation~\ref{#1}}
\def\1{\bm{1}}

\DeclareMathAlphabet{\mathsfit}{\encodingdefault}{\sfdefault}{m}{sl}
\SetMathAlphabet{\mathsfit}{bold}{\encodingdefault}{\sfdefault}{bx}{n}

\usepackage{hyperref}
\usepackage{url}
\usepackage{graphicx}

\title{How GPT Learns Layer by Layer}

\author{
Jason Du\textsuperscript{1}, Kelly Hong\textsuperscript{1}, Alishba Imran\textsuperscript{1}\thanks{Correspondence: alishbaimran@berkeley.edu}, 
Erfan Jahanparast\textsuperscript{1}, Mehdi Khfifi\textsuperscript{1}, Kaichun Qiao\textsuperscript{1}\thanks{Authors are listed alphabetically.} \\
\textsuperscript{1}Department of Electrical Engineering and Computer Sciences, University of California, Berkeley
}

 \iclrfinalcopy % Uncomment for camera-ready version, but NOT for submission.
\begin{document}

\maketitle

\begin{abstract}

Large Language Models (LLMs) excel at tasks like language processing, strategy games, and reasoning but struggle to build generalizable internal representations essential for adaptive decision-making in agents. For agents to effectively navigate complex environments, they must construct reliable world models. While LLMs perform well on specific benchmarks, they often fail to generalize, leading to brittle representations that limit their real-world effectiveness. Understanding how LLMs build internal world models is key to developing agents capable of consistent, adaptive behavior across tasks. We analyze OthelloGPT, a GPT-based model trained on Othello gameplay, as a controlled testbed for studying representation learning. Despite being trained solely on next-token prediction with random valid moves, OthelloGPT shows meaningful layer-wise progression in understanding board state and gameplay. Early layers capture static attributes like board edges, while deeper layers reflect dynamic tile changes. To interpret these representations, we compare Sparse Autoencoders (SAEs) with linear probes, finding that SAEs offer more robust, disentangled insights into compositional features, whereas linear probes mainly detect features useful for classification. We use SAEs to decode features related to tile color and tile stability, a previously unexamined feature that reflects complex gameplay concepts like board control and long-term planning. We study the progression of linear probe accuracy and tile color using both SAE's and linear probes to compare their effectiveness at capturing what the model is learning. Although we begin with a smaller language model, OthelloGPT, this study establishes a framework for understanding the internal representations learned by GPT models, transformers, and LLMs more broadly. Our code is publicly available: \href{https://github.com/ALT-JS/OthelloSAE}{GitHub Repository}.

% \href{https://drive.google.com/file/d/1TfE_6oiuJ0ZCrqqG72LBNUD97k4gQVot/view?usp=sharing}{Presentation}, and \href{https://drive.google.com/file/d/1WZCeCSrEphDNQXtWHeFvd1gEgMTSdcpO/view?usp=sharing}{Video}.

\end{abstract}

\section{Introduction}

Large language models (LLMs) exhibit remarkable capabilities across tasks like natural language processing, strategic games, and reasoning. However, their demonstrated proficiency in performing complex reasoning tasks raises an open question: Do LLMs construct accurate internal representations of the structures, rules, and patterns that underlie the data they are trained on, or are these representations incomplete and brittle? A recent survey \cite{10.1145/3641289} on LLM evaluation highlights the need for multidimensional assessment, emphasizing that current models often succeed in task-specific metrics but lack deeper generalization. Similarly, studies on co-temporal reasoning \cite{su2024living} reveal significant gaps in how LLMs handle concurrent or overlapping temporal events, further showing the inconsistency of their internal representations. For example, the Othello-GPT model \cite{li2023emergent} predicts legal moves and reconstructs game board states from sequence data which shows how LLMs can infer hidden states purely from sequential patterns. However, studies by \cite{Toshniwal_Wiseman_Livescu_Gimpel_2022} and \cite{vafa2024world} demonstrate that such models often fail to recover the full compositional structure of their domains. This limitation is particularly evident in navigation and deterministic finite automata tasks, where models exhibit brittleness under dynamic or adversarial conditions. In games like Othello or navigation tasks, while these models excel in next-token prediction, they fail to construct coherent and generalizable internal representations, reducing their reliability in downstream applications. The key motivating question becomes: How do GPT models construct their world models during training? Beyond flawed representations, LLMs often exhibit peculiar behaviors. For instance, \cite{vafa2024world} describes how models trained on synthetic data, such as random walks, sometimes build better world models than those trained on real-world data. This anomaly highlights how training conditions and evaluation methods heavily influence the quality of implicit representations. Understanding these implicit models requires dissecting the features learned during training and analyzing their role in shaping internal representations.

In this paper, we address these gaps using OthelloGPT as an interpretability testbed. By tracing the features learned at each layer of OthelloGPT, we aim to uncover how GPT-based models construct their world models during training. Our main contributions include:
\begin{itemize}
    \item \textbf{Comparison of Interpretability Methods:} We compare Sparse Autoencoders (SAEs) and linear probes to analyze the learned representations. Our experiments show that SAEs uncover more distinctive and disentangled features, particularly for compositional attributes, whereas linear probes primarily identify features that act as strong correlators to classification accuracy.
    \item \textbf{Layer-wise Feature Analysis:} We uncover a hierarchical progression in OthelloGPT’s learned features, where some layers encode general attributes like board shape and edges, while others shift to potentially capturing more dynamic aspects of gameplay, such as tile flips and changing board states towards the center tiles.
\end{itemize}

\subsection{Impact on Agents}
LLM-based agents depend heavily on their internal world models to infer the latent structures necessary for tasks like compositional reasoning and long-term planning. For instance, \cite{rothkopf2024proceduraladherenceinterpretabilityneurosymbolic} highlights that agents require consistent internal representations to maintain procedural adherence and interpretability over extended interactions. However, studies like \cite{lopez-latouche-etal-2023-generating} demonstrate that LLMs often fail in long-term planning, leading to behavior that becomes inconsistent with prior states, as observed in video game character dialogues where maintaining style and narrative coherence is crucial.  Similarly, \cite{doi:10.1073/pnas.2218523120} show that LLMs struggle with compositional reasoning, failing to synthesize structured responses in complex scenarios like causal inference tasks where even minor perturbations cause substantial deviations from correct or human-like reasoning. These limitations emphasize the importance of coherent and generalizable world models that go beyond next-token prediction accuracy, which remains a primary metric in most evaluations. Agent interpretability can be examined in multiple ways: 1. Point-in-time interpretability: Focuses on how individual tokens in a prompt or response influence an agent’s immediate decisions. 2. Procedural interpretability: Analyze how sequences of inputs shape an agent's long-term behavior. 3. Mechanistic interpretability: Investigates the underlying mechanisms and representations within the model that relate to agents' outputs. In our work, we focus on mechanistic interpretability as a fundamental approach to understanding how internal model structures translate into observable agent behaviors. Othello serves as an excellent testbed for studying neural networks due to its compositional reasoning requirements and layered decision-making processes. By using Othello-GPT, we investigate the features the model learns layer-by-layer to understand how Transformers \cite{vaswani2017attention} and GPT-like models \cite{brown2020language} construct their implicit world models. While this research aims to expand to LLMs, SLMs like Othello-GPT are valid test beds due to their similar architecture and lower computational cost. By studying the model behavior and interpretability techniques evaluation like linear probes and SAEs on these smaller models, we can devise frameworks transferable to larger models. 
\section{Related Works}

\subsection{Sparse Autoencoders (SAEs)}

Sparse Autoencoders (SAEs) offer a way forward by disentangling learned features into sparse, interpretable bases \cite{bricken2023monosemanticity}. SAEs transform high-dimensional neural activations into sparse, interpretable representations by minimizing reconstruction error. The objective function for an SAE is:
\begin{equation}
L(x, \hat{x}) = \left\Vert x - \hat{x}\right\Vert ^2 + \lambda \left\Vert h \right\Vert _1
\end{equation}

Here, $x$ represents the input, $\hat{x}$ the reconstructed input, $h$ the latent encoding, and $\lambda\left\Vert h \right\Vert _1$ enforces sparsity by penalizing the $L1$ norm of $h$. Unlike linear probes, which fit a classifier $z = Wh + b$ to activations $h$ and identify features accessible via linear separability, SAEs uncover more granular and disentangled representations which can be particularly useful for compositional or overlapping features. By penalizing activation magnitude, SAEs force the model to distribute its representations sparsely across the basis vectors which can make features more interpretable.

More recent studies \cite{quaisley2024}, \cite{robert2024} and \cite{girit2023} applied SAEs to Othello-GPT and demonstrated their ability to disentangle sparse features linked to board states and moves. However, prior work did not analyze features layer by layer or compare SAEs directly to linear probes across layers.

\subsection{Linear Probes}

Linear probes are a widely used tool for analyzing neural networks(\cite{alain2016understanding}, \cite{tenney2019bert} and \cite{belinkov2021probingclassifierspromisesshortcomings}). Linear probes effectively measure feature accessibility but do not disentangle overlapping or compositional features. Prior studies on OthelloGPT (\cite{hazineh2023linearlatentworldmodels}) revealed that deeper layers encode increasingly accurate board representations as they are linearly accessible for tasks like board state classification. However, these studies did not compare the interpretability of features learned via linear probes with those disentangled by SAEs.

The original Othello-GPT model was trained to predict legal moves from sequential game data and could reconstruct board states using nonlinear probes (\cite{li2023emergent}). Neel Nanda later showed that a linear probe could identify a “world representation” of the board state, shifting from traditional representations (e.g., “black’s turn” vs. “white’s turn”) to a model-derived interpretation (“my turn” vs. “their turn”) (\cite{nandaothello2024}). While effective at the classification task, linear probes do not disentangle features or capture compositional patterns, limiting their interpretability.

\subsection{Layer by Layer Analysis}
\cite{alain2016understanding} introduced the use of linear classifier probes to measure the representations encoded at each layer of a neural network. This analysis revealed that deeper layers progressively improve the linear separability of features as representations become more abstract and aligned with the model's task objective. Subsequent research on GPT models (\cite{tenney2019bert}, \cite{belinkov2021probingclassifierspromisesshortcomings}) extended these insights to transformers. For instance, studies on BERT showed that earlier layers encode syntactic information, while later layers capture semantic roles like coreference and entity types. This revealed a progression of linguistic abstraction across layers. However, probing classifiers, particularly linear probes, often struggle to disentangle overlapping or compositional features within the model's activations.

Our work extends Bengio’s findings by comparing the features detected by SAEs to those identified by linear probes, revealing how disentangled representations evolve with depth. Inspired by studies on concept depth (\cite{rothkopf2024proceduraladherenceinterpretabilityneurosymbolic}, \cite{jin2024exploring}), we explore how abstract and compositional features emerge layer by layer. By bridging interpretability with functionality, our approach provides a nuanced perspective on how OthelloGPT’s and generally LLMs internal world model supports strategic decision-making.

\section{Method}

\begin{figure}[h]
\begin{center}
    \includegraphics[width=0.9\linewidth]{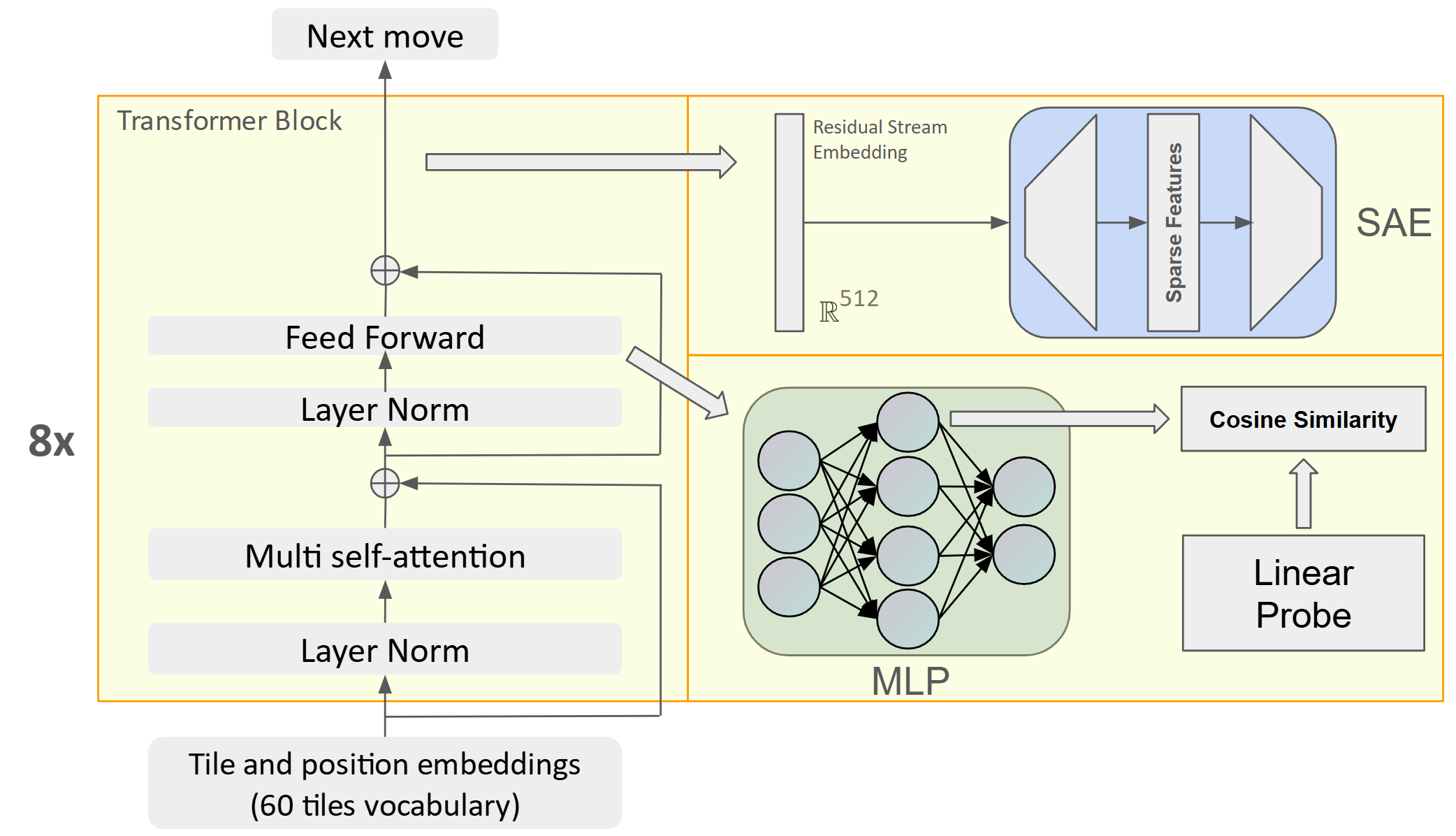}
\end{center} 
\caption{\textbf{Our work is divided into three parts.} The left side of the figure illustrates the architecture of OthelloGPT, designed to predict the next legal move in the game of Othello. The upper-right section shows how the Residual Stream from OthelloGPT is used as input to a SAE, enabling feature analysis through its sparse representations. The lower-right section presents a cosine similarity analysis between the parameters of individual neurons in the MLP layers of OthelloGPT and the linear probes we trained.}
\label{framework}
\end{figure}

\subsection{OthelloGPT}
We trained an 8-layer, decoder-only transformer model, Othello-GPT (as shown in Figure~\ref{framework}), with an 8-head attention mechanism and a residual stream dimension $d$ = 512, just like \cite{robert2024}. The model predicts the next token in random Othello game transcripts, treating games as sequences tokenized with a 66-word vocabulary (representing 64 board tiles, with a padding token and an end-of-sequence token). The OthelloGPT was trained on the Synthetic dataset mentioned in \cite{li2023emergent}. It contains out-of-distribution steps that are legal but sub-optimal, which conveys that our OthelloGPT training process has no long-term strategy involved. Formally, let the sequence of tokens for a game be $\mathbf{x} = (x_1, x_2, \dots, x_T)$, where $x_t$ is the token at timestep $t$. The model is trained to minimize the autoregressive next-token prediction loss:
$\mathcal{L}_{\text{token}} = - \frac{1}{T} \sum_{t=1}^T \log p(x_t \mid x_{<t})$, where $p(x_t \mid x_{<t}) = \text{softmax}(\mathbf{h}_t \mathbf{W}_{\text{output}})$, $\mathbf{h}_t$ is the hidden state of the model at timestep $t$, and $\mathbf{W}_{\text{output}}$ is the output projection matrix. Additionally, for our Sparse Autoencoder (SAE) experiments, we extracted residual stream embeddings from the intermediate layers of OthelloGPT and used these embeddings as inputs to the SAE.

\subsection{Linear Probe}
We trained linear probes across all layers of the 8-layer. Linear probes aim to extract specific semantic features or predict attributes from the model’s hidden states. Let the hidden states at layer $l$ of a decoder-only Transformer be $\mathbf{H}^{(l)} = [\mathbf{h}_1^{(l)}, \mathbf{h}_2^{(l)}, \dots, \mathbf{h}_n^{(l)}] \in \mathbb{R}^{n \times d}$, where $\mathbf{h}_i^{(l)} \in \mathbb{R}^d$ is the hidden representation for the $i$-th token, and $n$ is the sequence length. The linear probe is defined as a classifier: $g_{\phi}(\mathbf{h}) = \mathbf{W}^T \mathbf{h}$, where $\phi = \mathbf{W} \in \mathbb{R}^{d \times k}$ are the learnable parameters of the probe, and $k$ is the number of target classes. In our case, $k$ is 64 (similar to \cite{nandaothello2024}), which represents all the locations on the board. We also have trained three different probes under three modes respectively: empty, which means there are no game pieces on that location on the board; own, which means the game piece is my color on the board on that location; and enemy, which means the game piece is the opponent's color on the board on that location. For a given hidden state $\mathbf{h}$, the predicted probability distribution over the classes is: $\hat{\mathbf{y}} = \text{softmax}(\mathbf{W}^T \mathbf{h})$, and the predicted class is the one with the highest probability: $\hat{y} = \arg\max_j \hat{y}_j.$
The linear probe minimizes the cross-entropy loss function as follows:

% \(\mathcal{L}(\phi) = \frac{1}{N}\sum_{i = 1}^{N} \ell_{CE}(g_{\phi}(\mathbf{h}_i), y_{i})\), 

\begin{equation}
    \mathcal{L}(\phi) = \frac{1}{N}\sum_{i = 1}^{N} \ell_{CE}(g_{\phi}(\mathbf{h}_i), y_{i})
\end{equation}
where $N$ is the number of samples, $\mathbf{h}_i = f_{\theta}(\mathbf{x}_i)$ is the frozen hidden representation extracted by the Transformer $f_{\theta}$, $y_i$ is the corresponding target label, and the cross-entropy loss is defined as $\ell_{\text{CE}}(\hat{\mathbf{y}}, y) = -\log \hat{y}_y$, where $\hat{y}_y$ is the predicted probability for the correct class $y$. By training linear probes on hidden states from all layers, we evaluate the layer-wise encoding of semantic information in OthelloGPT.

\subsection{Tile color}
\label{subsec:tile_color_analysis}
We devised two methods, using linear probe and SAE, to analyze tile color and discover the robustness of features learned by OthelloGPT. 

\subsubsection{Linear Probe and Cosine Similarity}
To analyze the internal behavior of GPT, we employ the cosine similarity method for network analysis. Given two distinct feature $\mathbf{a}$ and $\mathbf{b}$, the cosine similarity function is: $similarity(\mathbf{a},\mathbf{b}) = \frac{\mathbf{a}\mathbf{b}}{\left\Vert \mathbf{a}\right\Vert \left\Vert \mathbf{b}\right\Vert}$. We focus on analyzing individual neurons in every layer. With a pre-trained linear probe for three modes, we calculate the cosine similarity between the MLP neurons and the probe, assessing each neuron's contribution to classification for specific tiles. We present findings primarily from the Encoding layer after observing similar results on the Encoding and Projection layers, where we compute the cosine similarity between MLP parameters and the layer-specific linear probe to quantify each neuron’s contribution to encoding tile color. Neurons exceeding a similarity threshold of 0.2 are counted for each tile, focusing on the “my color” probe.

\subsubsection{SAE}
To ensure the robustness of the features learned by Othello-GPT, we compared the top-performing features across 10 random initialization seeds of the SAEs. We extract sparse features from the residual stream embedding with shape $\mathbf{R}^{512}$ after feed forward in each transformer block. This validation ensures that the model is learning tile colors rather than artifacts from specific random states. For each seed, we identify the top 50 features with AUROC~$>$~0.7, which measure a feature's ability to classify tile states (empty, player’s piece, or opponent’s piece) by balancing the true positive and false positive rates. A higher AUROC value indicates stronger discriminative power. Aggregating results across all seeds, we tally the frequency of each board position appearing among these top features, as visualized in Figure~\ref{tile_color}.

\subsection{Tile Stability}
\label{subsec:stable_tile_analysis}
A tile is defined as stable if it cannot be flipped for the remainder of the game. For instance, corner tiles are inherently stable once placed. Stable tiles include corner tiles, edge tiles anchored to stable tiles, and interior tiles surrounded by stable tiles. Given a set of 104,000 board states (2,000 games x 52 board states per game), we computed binary stability maps for each state. Each board state consisted of 64 tiles, each encoded by its color (0 for empty, 1 for black, 2 for white).
Occupied tiles (indicated by 1 or 2), were marked as stable if it was 1) a corner tile ((0,0), (0,7), (7,0), or (7,7)) an edge tile directly adjacent to a stable tile or 3) an interior tile surrounded by 8 stable tiles (including top, bottom, left, right, and diagonally adjacent neighbors). This process yielded a stability map for each board state, where each tile was assigned a binary value: 1 if stable and 0 otherwise.

\textbf{Stability Feature Activations.} We analyzed feature activations across all 8 layers of our Othello-GPT model with a binary classification framework. Each board state was paired with its corresponding stability map to evaluate whether individual features reliably predicted tile stability. Features were considered active if their activation strength was $>$ 0.
For each feature and tile, we computed: true positives (active feature and stable tile), false positives (active feature, non-stable tile), true negatives (inactive feature, non-stable tile), and false negatives (inactive feature, stable tile). From these values, we calculated the F1-score and AUROC using standard metrics.

\textbf{Feature Analysis using AUROC Thresholds.}
To determine whether dominant features consistently encode stability across layers, we analyze feature activations with AUROC scores $>$ 0.8. We did the following analysis for each layer: 

1. Tile-level: For each tile, we computed the frequency of feature activations with AUROC scores exceeding the threshold. 

2. Feature-level: For each feature, we computed the frequency of its activations exceeding the threshold for that layer without regard to specific tile positions. We repeated this analysis for 2 different seeds to confirm our results.
\section{Experiments}
% We hypothesize that features are learned progressively across layers, with simpler features being encoded in the earlier layers and more complex features emerging in the later layers. To test this hypothesis, we conduct the following experiments:

% \begin{itemize}
%     \item \textbf{SAE vs Linear Probes for colors:} We evaluate the model’s ability to classify tile color. This is considered an easier task as it involves basic feature extraction without requiring an understanding of broader board dynamics. However, we still discovered differences between Linear Probes and SAEs where SAEs show a clearer, more distinct feature compared to Linear Probes.
%     \item \textbf{SAE for tile stability:} We analyze the model’s capacity to identify stable tiles (e.g., unflippable positions). This task is inherently harder as it requires the model to understand the state of the board and the game rules and positional dynamics. To our knowledge, tile stability has not been previously studied as a learned feature in GPT-based models.
% \end{itemize}

\subsection{Comparing SAEs vs Linear Probes}

\begin{figure}
    \centering
    \includegraphics[width=0.9\linewidth]{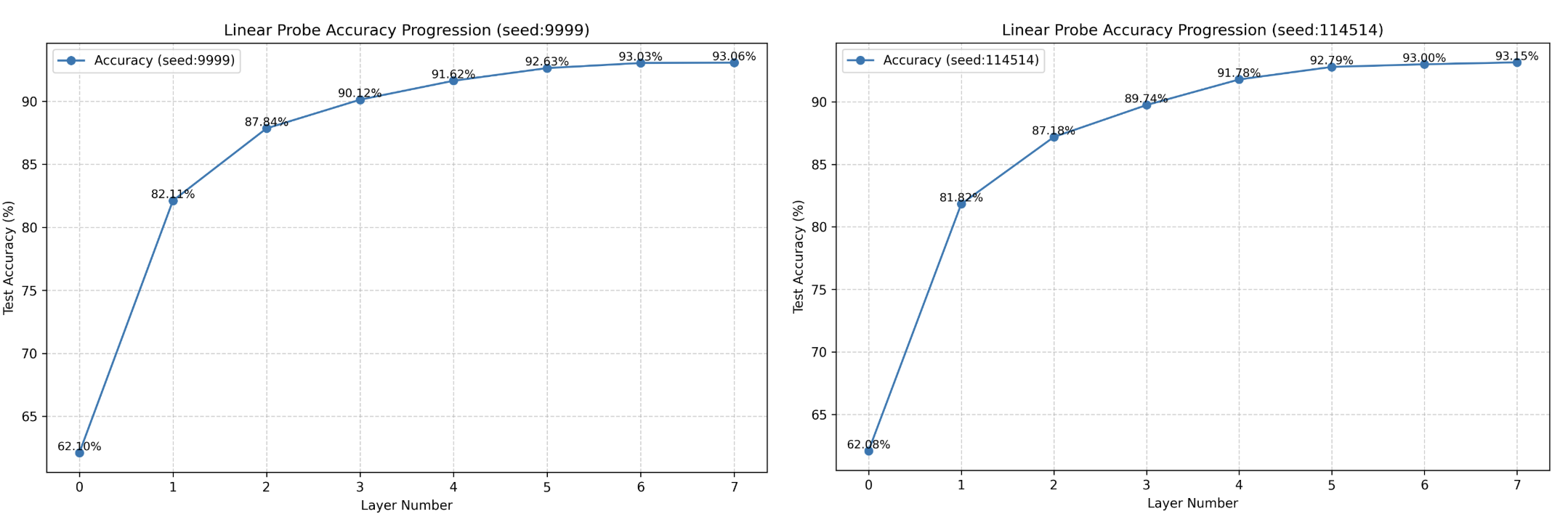}
    \caption{Linear probe accuracy for two seeds. The results demonstrate that linear probes effectively capture features that are good predictors of classification accuracy which increases over layers.}
    \label{fig:linear_progression}
\end{figure}

We show that linear probe accuracy increases across layers (Figure~\ref{fig:linear_progression}), suggesting the model learns stronger predictors for classification tasks. However, they fail to reveal distinct or compositional features per layer. SAEs address this by disentangling activations into sparse, interpretable bases, providing deeper insights into the features learned at each layer.

\subsection{Tile Color Across Layers}
% In the heatmaps shown in Figure~\ref{tile_color}, we observe a hierarchical progression in OthelloGPT’s learned features: earlier layers (1-2) focus on general attributes like edges and board shape, while deeper layers (4-7) increasingly activate central tiles, potentially reflecting more dynamic aspects of gameplay being learned such as tile flips and the changing board state. 

\begin{figure}[h]
\begin{center}
    \includegraphics[width=1.0\linewidth]{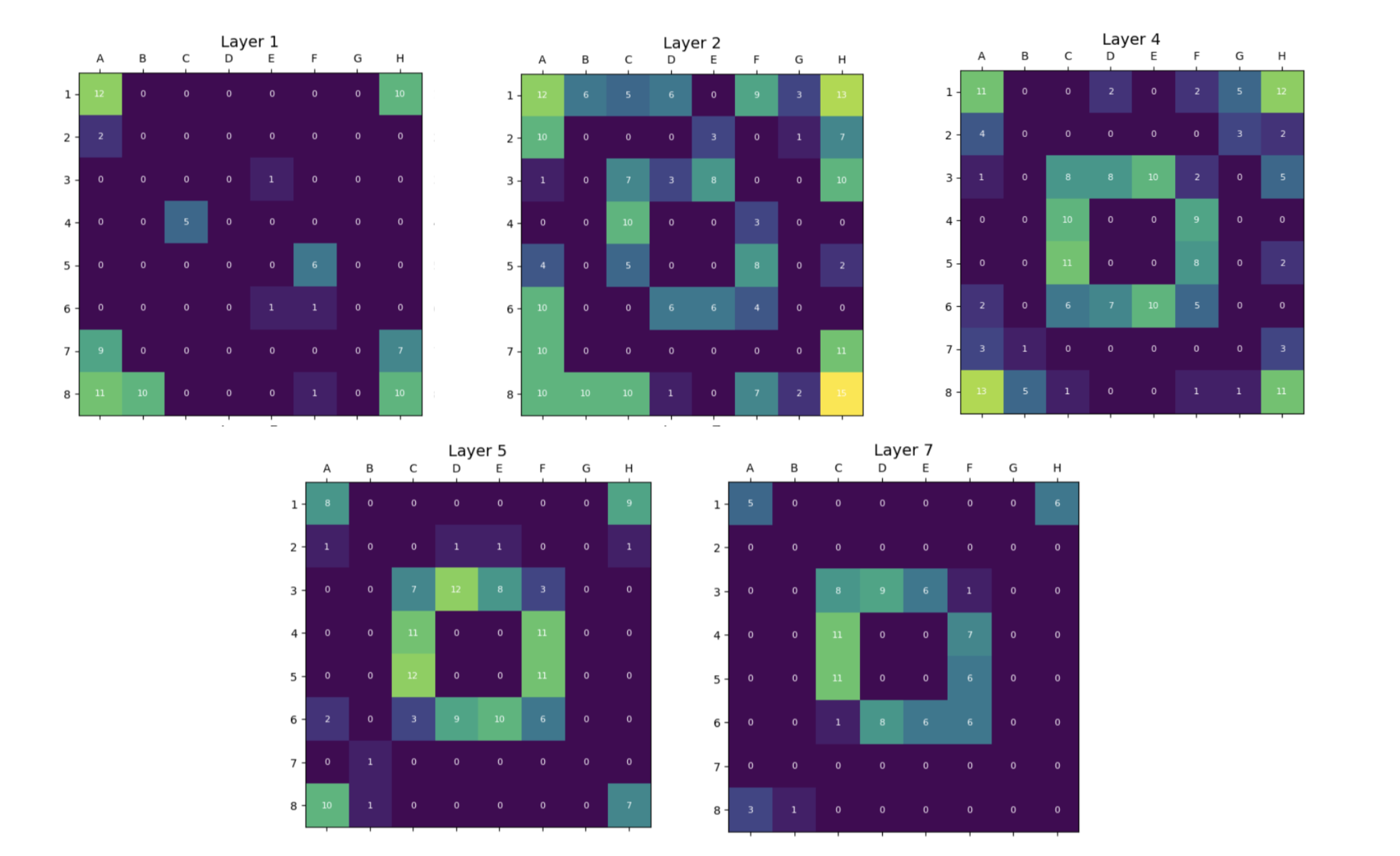}
\end{center} 
\caption{\textbf{SAE Tile color activation maps.} Showing frequency of tile color activations measured across 10 different seeds, as described in Section~\ref{subsec:tile_color_analysis}.}

\label{tile_color}
\end{figure}

\begin{figure}[h]
\begin{center}
    \includegraphics[width=0.9\linewidth]{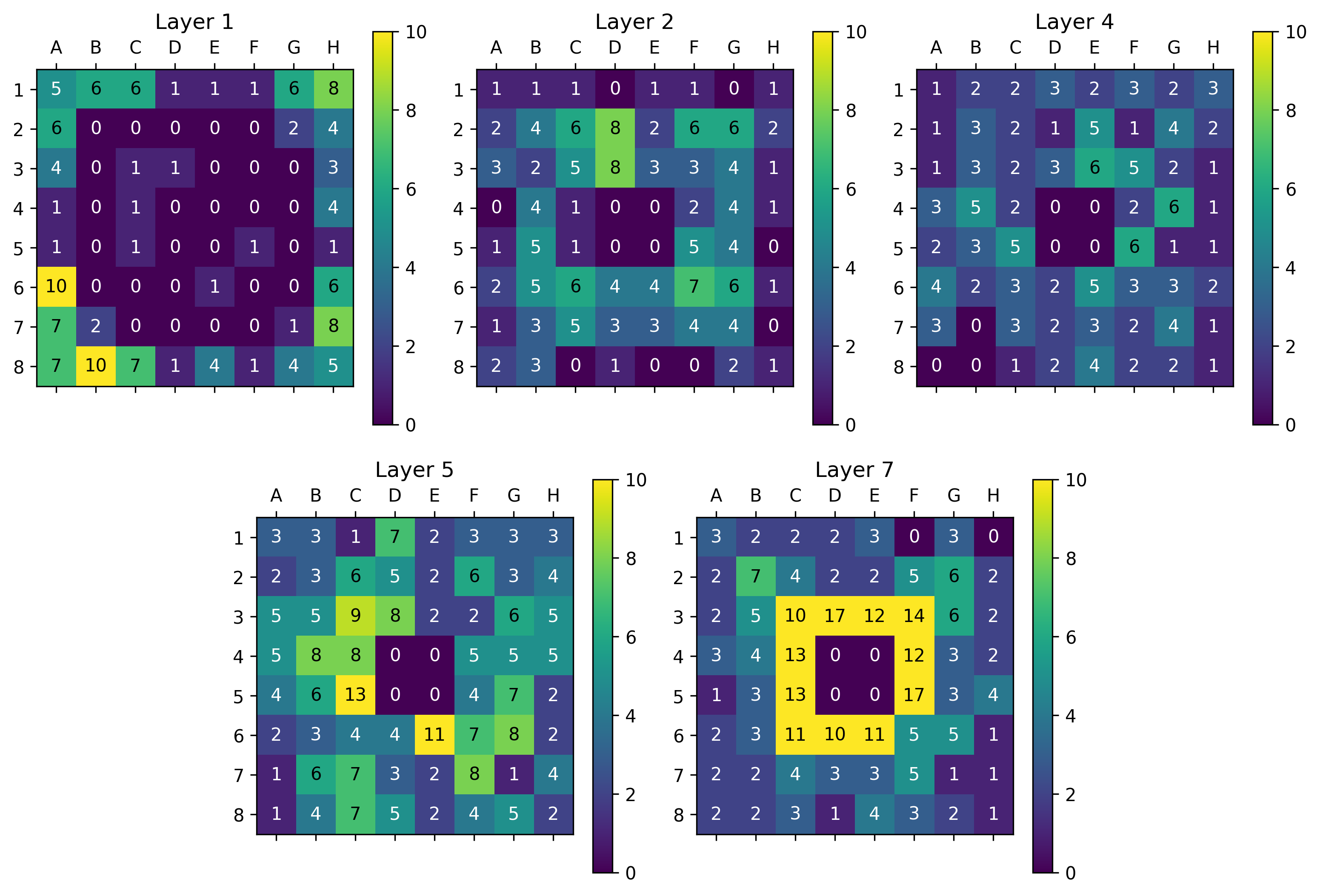}
\end{center} 
\caption{\textbf{Linear Probe tile color activation maps.} Showing the tile color activations measured across layers, as described in Section~\ref{subsec:tile_color_analysis}.}

\label{tile_color_linear_probe}
\end{figure}

Figure~\ref{tile_color} shows the features extracted by the SAEs, computed by identifying the most discriminative features using AUROC scores across multiple random seeds. The resulting heatmaps highlight positions with consistently high importance, such as edges and central tiles. In contrast, Figure~\ref{tile_color_linear_probe} visualizes the contributions of individual MLP neurons from linear probes to tile classifications, measured via cosine similarity.

Figure~\ref{tile_color} and Figure~\ref{tile_color_linear_probe} reveal distinct differences in how SAEs and linear probes learn features from the board. The SAE visualizations highlight clear and structured patterns, such as strong activations at corner and edge tiles in layer 1, indicating that the model captures the board's shape early on. As we progress to Layers 2 and 4, SAEs show more dynamic changes, with activations concentrated in the central tiles and along the edges. This suggests the SAEs are not only learning positional importance but could also capturing the evolving dynamics of central tiles, which tend to flip frequently as the game progresses. Importantly, these results are aggregated across 10 random seeds, demonstrating the robustness and consistency of SAEs in identifying meaningful features. In contrast, the linear probe visualizations show more dispersed activations across the board. While individual tiles are well-classified, the activations lack the clear structural patterns seen in SAEs.

\subsection{Tile Stability Across Layers}
\begin{figure}[h]
\begin{center}
    \includegraphics[width=0.9\linewidth]{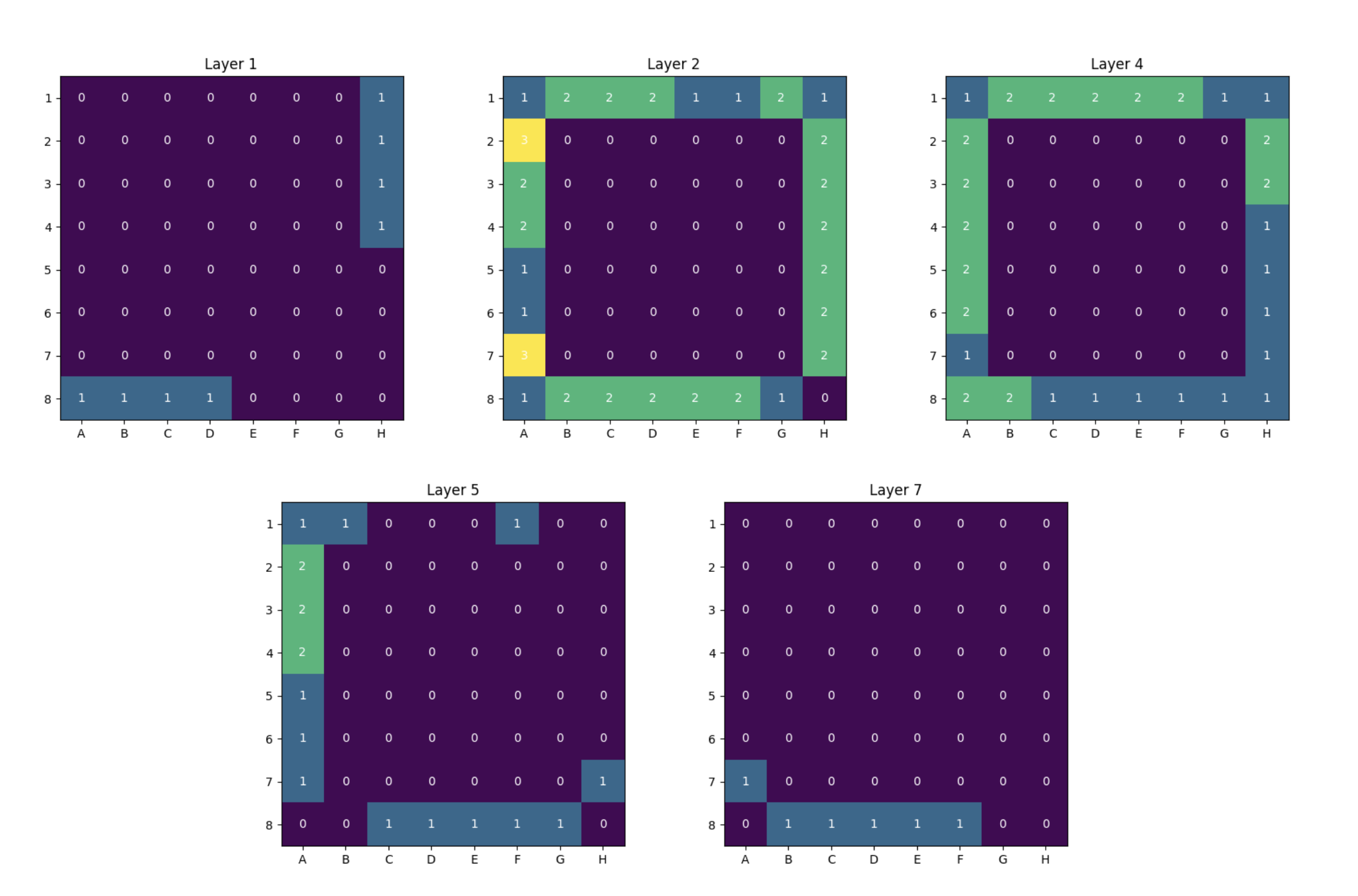}
\end{center} 
\caption{\textbf{Stability activation maps.} Computed for seed 1, as described in Section~\ref{subsec:stable_tile_analysis}. We provide results for another OthelloGPT model in Appendix Figure~\ref{stability_appendix} which shows a similar trend.}

\label{tile_stable}
\end{figure}

We notice the highest frequency of tile-feature activations in the intermediate layers (layers 2 through 4), which can be seen in Figure~\ref{tile_stable}. Earlier (layer 1) and later layers (layers 5 through 8) do not appear to learn stability, but rather they likely dedicate their representational capacity to other aspects of the board state. This layer-specific pattern is consistent with the notion that different depths in the model are dedicated to learning different concepts.

We further support this analysis through Table~\ref{tab:features_per_layer_1}, which reveal this same pattern across layers for distinct features. Features 349 and 108 for instance, show a strong pattern which suggests that they encode tile stability. Table~\ref{tab:top_auroc_layer_2_seed_1} shows the exact AUROC scores for tile-feature pairs in layer 2, which are clearly higher relative to other layers. We're able to disentangle and trace these feature patterns across layers through using SAEs, which is an advantage over using linear probes.

However, we must acknowledge that there is some variability across seeds as we can see in Table~\ref{tab:features_per_layer_1} and Table~\ref{tab:features_per_layer_seed2}. This variability raises the possibility that the features we interpret as "stability" may be a composition of related but more granular features, such as the presence of edge or corner tile configurations. These properties may jointly give rise to the notion of stability, without representing stability itself in isolation. Future work with additional seeds is necessary to fully investigate these distinctions. For example, fine-grained analyses of isolating and perturbing corner-detection features may shed light on the causal role of these subcomponents in producing emergent stability behaviors.

\section{Conclusion and Future Work}

% In this work, we analyzed the progression of learned features in OthelloGPT, uncovering a hierarchical structure where earlier layers capture general attributes like board edges and shape, while deeper layers focus on dynamic gameplay aspects, such as tile flips and evolving board states. By comparing SAEs and linear probes, we demonstrated that SAEs uncover more distinctive and disentangled features, particularly for compositional attributes, while linear probes primarily identify strong correlators for classification tasks. Through key experiments, we decoded features related to tile color and stability, establishing a framework for understanding how GPT-based models and transformers construct internal representations. Future work will use attribution analysis and automated interpretability methods \cite{bills2023language} to identify causal features influencing moves and better understand neuron-level representations.

%
In this work, we analyzed the progression of learned features in OthelloGPT, revealing a hierarchical structure in its internal representations. Specifically, we observe that different layers in OthelloGPT focus on distinct aspects of gameplay: some capture structural attributes like board shape and edges, while others appear to encode dynamic features such as tile flips and shifts in board state. By comparing the capabilities of sparse autoencoders (SAEs) and linear probes, we established that SAEs excel at uncovering more distinctive and disentangled features, particularly for compositional and interpretable attributes. In contrast, linear probes tend to highlight features that serve as strong correlators for classification tasks. Through these methods, we decoded features related to tile color and stability, offering a novel framework for understanding how GPT-based models and transformers construct and organize their internal representations.

Our findings suggest promising avenues for advancing model interpretability. Attribution analysis and automated interpretability methods, such as those proposed by \cite{bills2023language}, could be applied to identify causal features that directly influence move selection. This could lead to a deeper understanding of how individual neurons and attention heads contribute to decision-making processes. For future work, we want to expand our discovery in several ways. First, we can use fine-grained techniques to map the role of individual neurons in representing specific gameplay attributes or strategies, and see whether more neurons can behave similarly in patterns. We can also extend this analysis to other board game models and Large Language Models (LLMs) to determine whether hierarchical feature learning is consistent across tasks and model types. Finally, we believe that an important area of future work lies in comparing model-derived representations of Othello strategy with established human concepts of gameplay. This could involve mapping latent dimensions in the model to known strategic heuristics to understand where the model’s features align or conflict with human reasoning.

\section{Acknowledgments}
Authors are listed alphabetically. Alishba Imran, Kelly Hong, and Jason Du are co-leads of the study. Alishba Imran contributed to the design of experiments, writing, and editing the manuscript; Kelly Hong designed and evaluated the SAE tile stability analysis; and Jason Du focused on tile color and linear probe analysis.

\newpage



\bibliography{iclr2025_conference}
\bibliographystyle{iclr2025_conference}

\newpage
\appendix
\section{Appendix}
\setcounter{figure}{0}

\subsection{Features Activated for Stability}

\begin{figure}[h]
\begin{center}
    \includegraphics[width=0.9\linewidth]{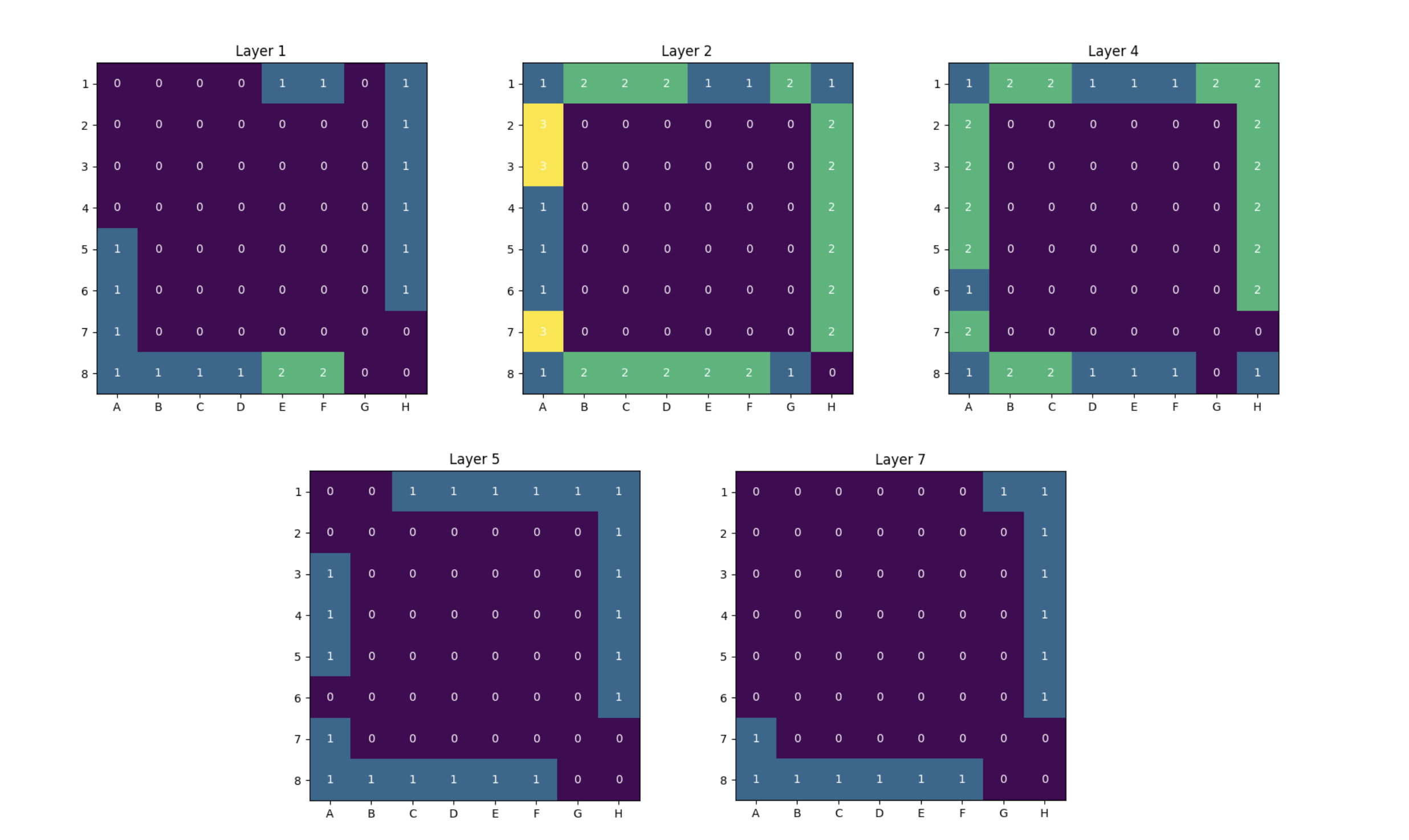}
\end{center} 
\caption{Tile Stability activation map (seed 2). }

\label{stability_appendix}
\end{figure}

\begin{table}[ht]
    \centering
    \resizebox{0.9\linewidth}{!}{%
    \begin{tabular}{|l|c|c|c|c|c|c|c|c|c|}
        \hline
        \textbf{Feature} & \textbf{Layer 1} & \textbf{Layer 2} & \textbf{Layer 3} & \textbf{Layer 4} & \textbf{Layer 5} & \textbf{Layer 6} & \textbf{Layer 7} & \textbf{Layer 8} & \textbf{Total Count} \\ \hline
        Feature 349 & 0 & 0 & 0 & 26 & 13 & 9 & 0 & 0 & 48 \\ \hline
        Feature 108 & 0 & 24 & 23 & 0 & 0 & 0 & 0 & 0 & 47 \\ \hline
        Feature 687 & 0 & 0 & 0 & 0 & 0 & 5 & 6 & 6 & 17 \\ \hline
        Feature 64  & 0 & 7 & 9 & 0 & 0 & 0 & 0 & 0 & 16 \\ \hline
        Feature 947 & 0 & 0 & 0 & 11 & 5 & 0 & 0 & 0 & 16 \\ \hline
        Feature 850 & 0 & 7 & 7 & 0 & 0 & 0 & 0 & 0 & 14 \\ \hline
        Feature 917 & 0 & 7 & 0 & 0 & 0 & 0 & 0 & 0 & 7 \\ \hline
        Feature 121 & 0 & 0 & 0 & 7 & 0 & 0 & 0 & 0 & 7 \\ \hline
        Feature 214 & 4 & 0 & 0 & 0 & 0 & 0 & 0 & 0 & 4 \\ \hline
        Feature 311 & 4 & 0 & 0 & 0 & 0 & 0 & 0 & 0 & 4 \\ \hline
        Feature 629 & 0 & 0 & 3 & 0 & 0 & 0 & 0 & 0 & 3 \\ \hline
        Feature 706 & 0 & 0 & 0 & 3 & 0 & 0 & 0 & 0 & 3 \\ \hline
        Feature 689 & 0 & 0 & 0 & 0 & 0 & 2 & 0 & 0 & 2 \\ \hline
        Feature 921 & 0 & 0 & 1 & 0 & 0 & 0 & 0 & 0 & 1 \\ \hline
        Feature 20  & 0 & 0 & 1 & 0 & 0 & 0 & 0 & 0 & 1 \\ \hline
        Feature 423 & 0 & 0 & 1 & 0 & 0 & 0 & 0 & 0 & 1 \\ \hline
        Feature 385 & 0 & 0 & 1 & 0 & 0 & 0 & 0 & 0 & 1 \\ \hline
        Feature 690 & 0 & 0 & 0 & 0 & 0 & 0 & 0 & 0 & 1 \\ \hline
        Feature 691 & 0 & 0 & 0 & 0 & 0 & 0 & 0 & 0 & 1 \\ \hline
        Feature 678 & 0 & 0 & 0 & 0 & 0 & 0 & 0 & 0 & 1 \\ \hline
    \end{tabular}%
    }
    \caption{Features activated per layer for stability (seed 1)}
    \label{tab:features_per_layer_1}
\end{table}

\begin{table}[ht]
    \centering
    \resizebox{0.9\linewidth}{!}{%
    \begin{tabular}{|l|c|c|c|c|c|c|c|c|c|}
        \hline
        \textbf{Feature} & \textbf{Layer 1} & \textbf{Layer 2} & \textbf{Layer 3} & \textbf{Layer 4} & \textbf{Layer 5} & \textbf{Layer 6} & \textbf{Layer 7} & \textbf{Layer 8} & \textbf{Total Count} \\ \hline
        Feature 90  & 0 & 0 & 11 & 20 & 14 & 0 & 0 & 0 & 45 \\ \hline
        Feature 67  & 0 & 0 & 0  & 7  & 7  & 7 & 7 & 6 & 34 \\ \hline
        Feature 403 & 0 & 24 & 0  & 0  & 0  & 0 & 0 & 0 & 24 \\ \hline
        Feature 727 & 0 & 7  & 7  & 7  & 0  & 0 & 0 & 0 & 21 \\ \hline
        Feature 369 & 0 & 0  & 0  & 0  & 5  & 7 & 7 & 7 & 19 \\ \hline
        Feature 50  & 0 & 0  & 0  & 7  & 0  & 0 & 0 & 0 & 14 \\ \hline
        Feature 629 & 8 & 0  & 0  & 0  & 0  & 0 & 0 & 0 & 8  \\ \hline
        Feature 76  & 7 & 0  & 0  & 0  & 0  & 0 & 0 & 0 & 7  \\ \hline
        Feature 1016 & 0 & 0 & 7  & 0  & 0  & 0 & 0 & 0 & 7  \\ \hline
        Feature 373 & 0 & 0  & 6  & 0  & 0  & 0 & 0 & 0 & 6  \\ \hline
        Feature 412 & 4 & 0  & 0  & 0  & 0  & 0 & 0 & 0 & 4  \\ \hline
        Feature 130 & 0 & 2  & 0  & 0  & 0  & 0 & 0 & 0 & 2  \\ \hline
        Feature 196 & 0 & 0  & 0  & 1  & 0  & 0 & 0 & 0 & 1  \\ \hline
        Feature 233 & 0 & 0  & 0  & 1  & 0  & 0 & 0 & 0 & 1  \\ \hline
        Feature 158 & 0 & 0  & 0  & 1  & 0  & 0 & 0 & 0 & 1  \\ \hline
        Feature 496 & 0 & 0  & 1  & 0  & 0  & 0 & 0 & 0 & 1  \\ \hline
        Feature 647 & 0 & 0  & 0  & 0  & 0  & 0 & 0 & 1 & 1  \\ \hline
        Feature 270 & 0 & 0  & 0  & 0  & 0  & 0 & 0 & 1 & 1  \\ \hline
        Feature 263 & 0 & 0  & 0  & 0  & 0  & 0 & 0 & 1 & 1  \\ \hline
        Feature 514 & 0 & 0  & 0  & 0  & 0  & 0 & 0 & 1 & 1  \\ \hline
    \end{tabular}%
    }
    \caption{Features activated per layer for stability (seed 2)}
    \label{tab:features_per_layer_seed2}
\end{table}

\begin{table}[ht]
    \centering
    {%
    \begin{tabular}{|c|c|c|}
        \hline
        \textbf{Feature} & \textbf{Tile Number} & \textbf{AUROC} \\ \hline
        214 & 7  & 0.8814 \\ \hline
        311 & 56 & 0.8731 \\ \hline
        214 & 15 & 0.8412 \\ \hline
        311 & 57 & 0.8344 \\ \hline
        214 & 23 & 0.8246 \\ \hline
        311 & 58 & 0.8174 \\ \hline
        214 & 31 & 0.8105 \\ \hline
        311 & 59 & 0.8020 \\ \hline
    \end{tabular}%
    }
    \caption{Top Feature-Tile AUROC Scores (Layer 1, Seed 1)}
    \label{tab:top_auroc_layer_1_seed_1}
\end{table}

\begin{table}[ht]
    \centering
    {%
    \begin{tabular}{|c|c|c|}
        \hline
        \textbf{Feature} & \textbf{Tile Number} & \textbf{AUROC} \\ \hline
        850 & 56 & 0.9504 \\ \hline
        917 & 7  & 0.9476 \\ \hline
        917 & 15 & 0.9074 \\ \hline
        850 & 57 & 0.9052 \\ \hline
        64  & 0  & 0.8936 \\ \hline
        917 & 23 & 0.8907 \\ \hline
        850 & 58 & 0.8859 \\ \hline
        917 & 31 & 0.8773 \\ \hline
        850 & 59 & 0.8696 \\ \hline
        917 & 39 & 0.8642 \\ \hline
        108 & 61 & 0.8595 \\ \hline
        917 & 6  & 0.8592 \\ \hline
        385 & 55 & 0.8584 \\ \hline
        108 & 40 & 0.8569 \\ \hline
        850 & 60 & 0.8561 \\ \hline
        850 & 48 & 0.8551 \\ \hline
        108 & 32 & 0.8545 \\ \hline
        108 & 60 & 0.8535 \\ \hline
        108 & 5  & 0.8529 \\ \hline
        917 & 47 & 0.8506 \\ \hline
    \end{tabular}%
    }
    \caption{Top Feature-Tile AUROC Scores (Layer 2, Seed 1)}
    \label{tab:top_auroc_layer_2_seed_1}
\end{table}

\begin{table}[ht]
    \centering
    {%
    \begin{tabular}{|c|c|c|}
        \hline
        \textbf{Feature} & \textbf{Tile Number} & \textbf{AUROC} \\ \hline
        947 & 0  & 0.8999 \\ \hline
        947 & 8  & 0.8767 \\ \hline
        947 & 1  & 0.8682 \\ \hline
        947 & 16 & 0.8568 \\ \hline
        349 & 55 & 0.8510 \\ \hline
        349 & 48 & 0.8446 \\ \hline
        349 & 61 & 0.8444 \\ \hline
        349 & 32 & 0.8443 \\ \hline
        349 & 59 & 0.8437 \\ \hline
        947 & 2  & 0.8437 \\ \hline
        947 & 24 & 0.8435 \\ \hline
        706 & 7  & 0.8427 \\ \hline
        349 & 60 & 0.8419 \\ \hline
        349 & 40 & 0.8416 \\ \hline
        349 & 62 & 0.8354 \\ \hline
        349 & 24 & 0.8347 \\ \hline
        947 & 3  & 0.8307 \\ \hline
        349 & 58 & 0.8303 \\ \hline
        349 & 5  & 0.8300 \\ \hline
        349 & 47 & 0.8296 \\ \hline
    \end{tabular}%
    }
    \caption{Top Feature-Tile AUROC Scores (Layer 4, Seed 1)}
    \label{tab:top_auroc_layer_4_seed_1}
\end{table}

\begin{table}[ht]
    \centering
    {%
    \begin{tabular}{|c|c|c|}
        \hline
        \textbf{Feature} & \textbf{Tile Number} & \textbf{AUROC} \\ \hline
        349 & 32 & 0.8532 \\ \hline
        349 & 40 & 0.8524 \\ \hline
        947 & 0  & 0.8483 \\ \hline
        349 & 61 & 0.8368 \\ \hline
        947 & 8  & 0.8367 \\ \hline
        349 & 48 & 0.8351 \\ \hline
        349 & 24 & 0.8332 \\ \hline
        349 & 59 & 0.8291 \\ \hline
        349 & 60 & 0.8288 \\ \hline
        349 & 16 & 0.8233 \\ \hline
        947 & 16 & 0.8227 \\ \hline
        947 & 1  & 0.8189 \\ \hline
        947 & 24 & 0.8094 \\ \hline
        349 & 55 & 0.8067 \\ \hline
        349 & 58 & 0.8056 \\ \hline
        349 & 62 & 0.8054 \\ \hline
        349 & 8  & 0.8047 \\ \hline
        349 & 5  & 0.8042 \\ \hline
    \end{tabular}%
    }
    \caption{Top Feature-Tile AUROC Scores (Layer 5, Seed 1)}
    \label{tab:top_auroc_layer_5_seed_1}
\end{table}

\begin{table}[ht]
    \centering
    {%
    \begin{tabular}{|c|c|c|}
        \hline
        \textbf{Feature} & \textbf{Tile Number} & \textbf{AUROC} \\ \hline
        687 & 60 & 0.8495 \\ \hline
        687 & 59 & 0.8488 \\ \hline
        687 & 61 & 0.8439 \\ \hline
        687 & 48 & 0.8438 \\ \hline
        687 & 58 & 0.8366 \\ \hline
        687 & 57 & 0.8247 \\ \hline
    \end{tabular}%
    }
    \caption{Top Feature-Tile AUROC Scores (Layer 7, Seed 1)}
    \label{tab:top_auroc_layer_7_seed_1}
\end{table}

\end{document}